\pdfoutput=1
\documentclass[11pt,a4paper]{article}
\usepackage[hyperref]{naaclhlt2019}
\usepackage{times}
\usepackage{latexsym}
\usepackage{subcaption}
\usepackage{dblfloatfix}

\usepackage{enumitem}
\usepackage{amsmath}
\usepackage{graphicx}

\newcommand{\nomiclogosize}{1.4em}
\DeclareRobustCommand{\nomicN}{%
  \raisebox{-0.3em}{%
    \includegraphics[height=\nomiclogosize,trim=15 10 20 10,clip]{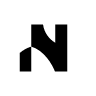}%
  }%
}

\usepackage{tikz}
\usepackage{caption}
\usepackage{pgfplots}
\usepackage{makecell}
\usepackage{booktabs, multirow, tabularx} 
\setlist[enumerate]{label*=\arabic*.}
\usepackage{url}

\aclfinalcopy
\title{AEC-Bench: A Multimodal Benchmark for Agentic Systems in Architecture, Engineering, and Construction}

\author{
Harsh Mankodiya\\
\texttt{\small harsh@nomic.ai}
\And
Chase Gallik\\
\texttt{\small chase@nomic.ai}
\AND
Theodoros Galanos\\
\texttt{\small theodoros.galanos@aurecongroup.com}
\And
Andriy Mulyar\\
\texttt{\small andriy@nomic.ai}
}

\begin{document}
\maketitle

\begin{abstract}
The AEC-Bench is a multimodal benchmark for evaluating agentic systems on real-world tasks in the Architecture, Engineering, and Construction (AEC) domain. The benchmark covers tasks requiring drawing understanding, cross-sheet reasoning, and construction project-level coordination. This report describes the benchmark motivation, dataset taxonomy, evaluation protocol, and baseline results across several domain-specific foundation model harnesses. We use AEC-Bench to identify consistent tools and harness design techniques that uniformly improve performance across foundation models in their own base harnesses, such as Claude Code and Codex. We openly release our benchmark dataset, agent harness, and evaluation code for full replicability at \href{https://github.com/nomic-ai/aec-bench}{https://github.com/nomic-ai/aec-bench} under an Apache 2 license.
\end{abstract}

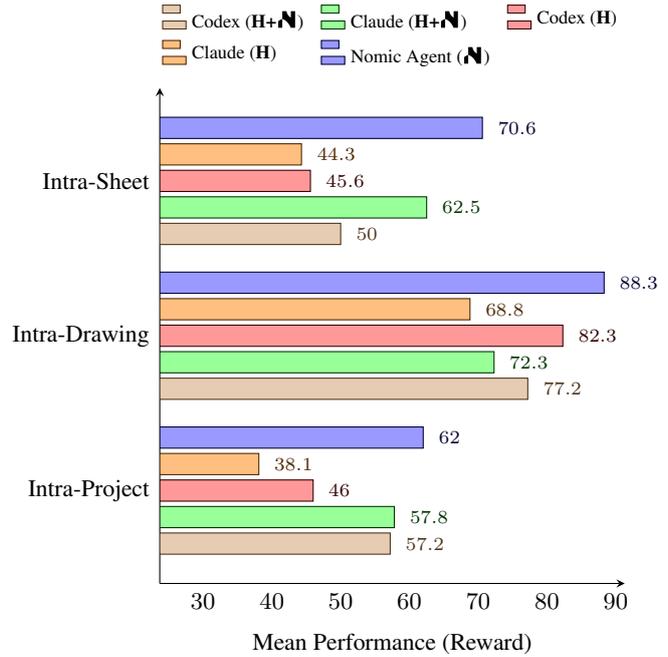
\begin{figure}[!t]
\hspace*{-0.05\textwidth}
\captionsetup{justification=raggedright,singlelinecheck=false}
\centering
\begin{tikzpicture}
  \begin{axis}[
    width=\linewidth,
    height=0.33\textheight,
    xbar,
    bar width=8pt,
    y axis line style = { opacity = 1 },
    axis y line = left,
    axis x line       = bottom,
    tickwidth         = 0pt,
    xmin = 25,
    xmax = 90,
    xtick distance=10,
    xlabel={Mean Performance (Reward)},
    xlabel style={font=\small},
    tick label style={font=\small},
    symbolic y coords = {Intra-Project, Intra-Drawing, Intra-Sheet},
    ytick=data,
    enlarge y limits  = 0.3,
    enlarge x limits  = 0.02,
    nodes near coords,
    every node near coord/.append style={font=\scriptsize, xshift=2pt},
    legend columns=3,
    legend style={
      font=\scriptsize,
      at={(0.5,1.02)},
      anchor=south,
      draw=none,
      /tikz/every even column/.append style={column sep=0.4em},
      cells={anchor=west},
    },
  ]

  \addplot[brown!40!black,fill=brown!40!white] coordinates {
    (57.2,Intra-Project)
    (77.2,Intra-Drawing)
    (50.0,Intra-Sheet)
  };

  \addplot[green!20!black,fill=green!40!white] coordinates {
    (57.8,Intra-Project)
    (72.3,Intra-Drawing)
    (62.5,Intra-Sheet)
  };

  \addplot[red!20!black,fill=red!40!white] coordinates {
    (46.0,Intra-Project)
    (82.3,Intra-Drawing)
    (45.6,Intra-Sheet)
  };

  \addplot[orange!30!black,fill=orange!50!white] coordinates {
    (38.1,Intra-Project)
    (68.8,Intra-Drawing)
    (44.3,Intra-Sheet)
  };

  \addplot[blue!20!black,fill=blue!40!white] coordinates {
    (62.0,Intra-Project)
    (88.3,Intra-Drawing)
    (70.6,Intra-Sheet)
  };

  \legend{Codex (\textbf{H+\nomicN}), Claude (\textbf{H+\nomicN}), Codex (\textbf{H}), Claude (\textbf{H}), Nomic Agent (\nomicN)}

      \end{axis}
\end{tikzpicture}

\caption{\textbf{Mean reward by AEC-Bench category.} {AEC-Bench mean reward by task, model, and setup. Rows are grouped by benchmark category and task; each model has two columns: \textbf{H} (base agent harness) and \textbf{H+\nomicN} (base agent harness augmented with Nomic tools and models).}}
\end{figure}

\begin{figure*}[t]
\captionsetup{justification=raggedright,singlelinecheck=false}
\centering
\includegraphics[width=\textwidth]{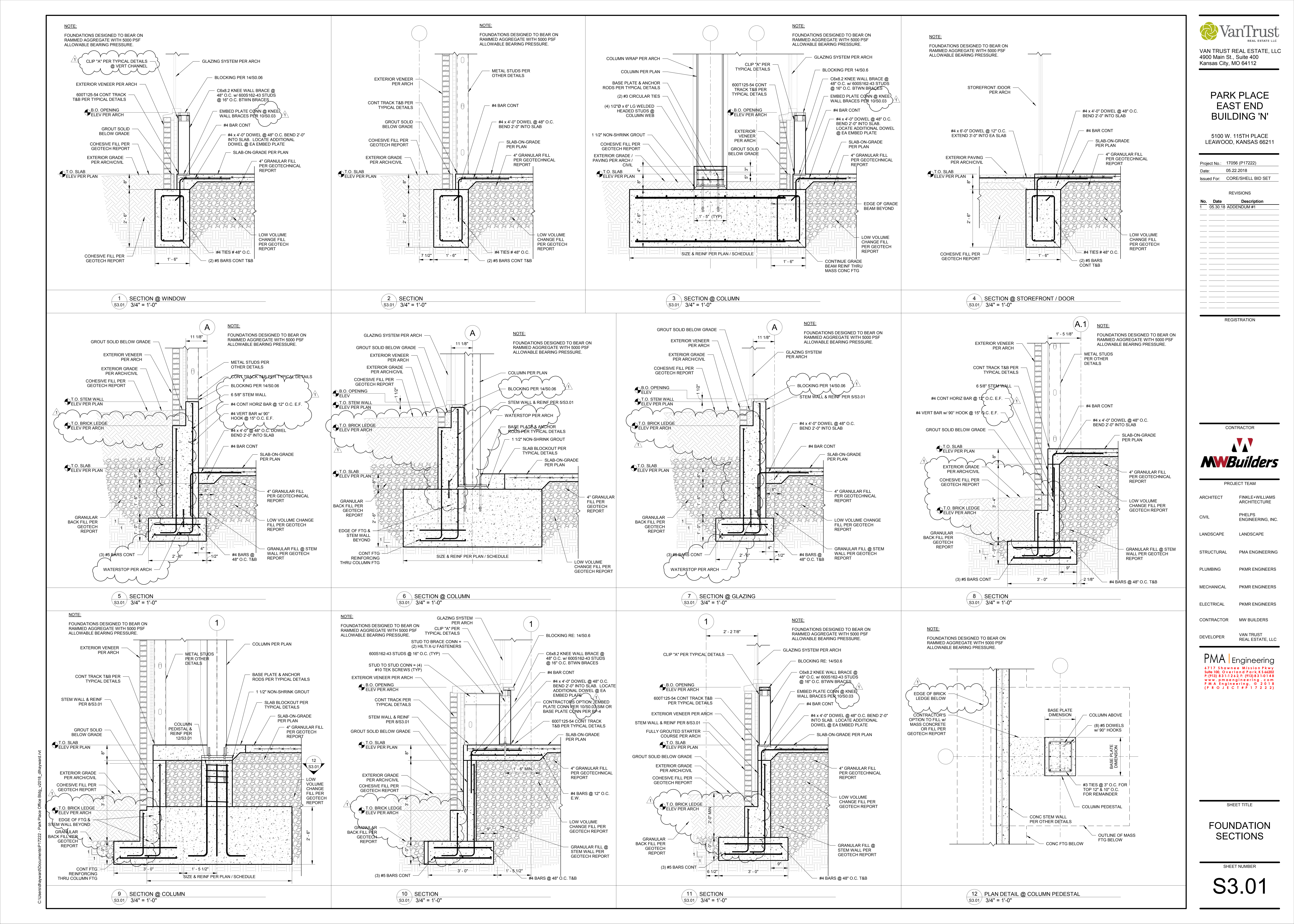}
\caption{\textbf{Example of a complex construction drawing PDF page.} A visually dense drawing sheet with tightly packed annotations, callouts, and linework typical of real construction coordination artifacts.}
\label{fig:complex-diagram}
\end{figure*}

\section{Introduction}

Foundation models equipped with coding-agent harnesses demonstrate strong capabilities in software engineering workflows, where agents can search repositories, edit code, and verify outputs through tool use. These systems rely on structured, verifiable environments and well-defined execution primitives, which enable reliable multi-step reasoning and self-correction. As a result, such harnesses are considered a recipe for building capable agents across new data and task domains.

Tasks across architecture, engineering, and construction require multimodal understanding, presenting unique challenges to building reliable agents. For example, a single construction drawing sheet, Figure~\ref{fig:complex-diagram}, contains tightly packed annotations, callouts, line work, and cross-references that require visual reasoning. Agents in construction require access to information that is distributed across highly multimodal documents, resulting in the failure of tools like text search. Standard tools in popular agent harnesses, such as text extraction, flatten spatial structure, while vision-based tools lack the quality needed for reliable geometric reasoning. As a result, agents applied to construction tasks often retrieve incomplete or incorrect context, leading to compounding errors. To study these limitations, we built AEC-Bench, a multimodal benchmark to evaluate agentic systems in real-world architecture, engineering, and construction workflows. The benchmark consists of a scaffolded collection of tasks drawn from coordination practices, including intra-sheet review, cross-sheet navigation, and project-level document alignment. We evaluate state-of-the-art coding-agent harnesses on this benchmark to characterize where they succeed, where they fail, and how domain-specific tooling affects performance.

We release the benchmark dataset, agent harnesses, and evaluation code to support reproducible research on multimodal agents operating in complex document environments.

\subsection{Contributions}
Our main contributions are:
\begin{itemize}[leftmargin=*]
\item We introduce AEC-Bench, a multimodal benchmark for agentic systems in engineering and construction, with a taxonomy spanning intra-sheet, intra-drawing, and intra-project reasoning.
\item We utilize domain-experts to curate a benchmark with 196 instances across 9 task families, grounded in construction coordination workflows and paired with automated evaluation.
\item We show that general-purpose coding-agents partially generalize to AEC tasks but fail on visual grounding, exhaustive traversal, and cross-document coordination.
\item We demonstrate that domain specific tools like parsing improve performance on some retrieval-sensitive tasks, but further work is required for grounding and judgment-heavy tasks.
\end{itemize}

\subsection{Related Work}

The AEC-Bench has been developed to help identify and characterize the boundary points of agent performance in a high-value construction coordination task. These coordination tasks require agents to demonstrate multimodal document understanding and context engineering capabilities over long horizons. Recent benchmarks for agentic systems demonstrate that tool use, scaffolding, and verifier design strongly influence measured capabilities. For example, SWE-bench \cite{jimenez2024swebench} evaluates agents in software engineering environments where models must navigate repositories, edit files, and verify outputs through automated tests. Related efforts such as GAIA \cite{mialon2023gaiabenchmarkgeneralai} and BFCL \cite{patil2025the} evaluate the ability of agents to plan multi-step actions and invoke external tools or function calls. Related agent evaluation environments such as AgentBench \cite{liu2024agentbench} and WebArena \cite{zhou2023webarena} similarly require interaction with complex textual artifacts and external resources during multi-step reasoning, highlighting that agent performance depends not only on the underlying model but also on the surrounding execution environment and evaluation protocol.

Another line of work focuses on multimodal reasoning and document understanding. Early benchmarks such as DocVQA and InfographicVQA \cite{mathew2021docvqadatasetvqadocument} evaluate visual question answering over document images containing tables, forms, and graphical elements. Subsequent datasets such as DUDE \cite{vanlandeghem2023documentunderstandingdatasetevaluation} extend this setting to long documents and multi-page contexts. More recent benchmarks explore long-context and multi-document reasoning, including MMLongBench-Doc \cite{ma2024mmlongbenchdocbenchmarkinglongcontextdocument}, LongDocURL \cite{deng2025longdocurlcomprehensivemultimodallong}, M-LongDoc \cite{chia-etal-2025-longdoc}, and M3DocVQA \cite{cho2025m3docvqa}, which evaluate models on tasks requiring navigation across pages and integration of information from multiple document segments. Complementary multimodal reasoning benchmarks such as MMMU \cite{yue2024mmmumassivemultidisciplinemultimodal} and ChartQA \cite{masry-etal-2022-chartqa} study reasoning over visually rich content containing charts, diagrams, and embedded text.

A smaller body of work explores agent interactions with document collections. Systems such as DocPrompting \cite{docprompting2024} and retrieval-based document agents such as ViDoRe \cite{loison2026vidorev3comprehensiveevaluation} evaluate models that iteratively retrieve, read, and reason over document corpora. While this line of work highlights the importance of agent scaffolding for interacting with documents and external knowledge sources, it primarily focuses on retrieval and question answering over textual content.

Recent AEC-specific benchmarks further motivate the need to distinguish between domain knowledge evaluation, drawing perception, and workflow-level agent performance. \cite{Liang_2026} introduces a hierarchical benchmark for LLM knowledge evaluation in AEC. While it broadens domain-specific evaluation beyond narrow exam-style settings, its primary focus is measuring AEC knowledge and cognitive proficiency rather than evaluating tool-using agents operating over multimodal project artifacts. AECV-Bench \cite{kondratenko2026aecvbenchbenchmarkingmultimodalmodels}, instead, focuses on multimodal understanding of architectural and engineering drawings, evaluating capabilities such as OCR, counting, spatial reasoning, and drawing-grounded question answering over floor plans and related artifacts. Its findings show that current multimodal models can function as document assistants but still lack robust drawing literacy, especially for symbol-centric understanding. Recent work on agent benchmarking in real engineering environments further reinforces the importance of evaluating systems within realistic execution contexts rather than isolated capability tests \cite{galanos2026benchmarking_agents_engineering}. In contrast, our AEC-Bench evaluates agents in realistic workflows where success depends not only on perception or domain knowledge but also on retrieval, cross-sheet navigation, cross-document reasoning, and structured reporting under an execution harness. This places our benchmark closer to workflow evaluation for agentic systems than to static knowledge testing or document visual QA.

\subsection{Construction Coordination}

Physical assets such as bridges, water treatment facilities, and office buildings are designed, engineered, and constructed by skilled professionals whose expertise is developed through years of post-graduate training. While designers and engineers perform preliminary work in 3D modeling software such as Revit and AutoCAD, most stages of design development, coordination, and delivery are coordinated through 2D construction drawing sets and related documents. A drawing set and its related documents, such as project specifications, communicate the requirements and design intent for constructing a physical asset. In practical terms, a drawing set is a multi-page, multimodal instruction package in which meaning is conveyed through structured visual and textual elements such as plans, details, callouts, notes, and title blocks.

Coordination failures between architects, engineers, and construction teams are often the main drivers of scheduling delays and budget overruns when designing and building physical assets. During pre-construction, design, engineering, and handoff to the construction team, many delays arise from inconsistencies introduced while authoring and revising drawing sets and project documents. In response, industry teams rely on standardized review and coordination workflows that demand deep professional experience and multimodal reasoning. These workflow characteristics make AEC a strong setting for evaluating agentic systems that must interpret multimodal inputs, reason between documents, and produce structured findings under operational and physical constraints. In practice, professional review workflows decompose into distinct reasoning tasks that vary in the amount of document context required. Some checks can be performed by inspecting a single sheet (page of a drawing set), while others require tracing references across drawing sets or coordinating information across multiple project artifacts. To reflect these differing context requirements, we organize benchmark tasks using a taxonomy based on the scope of context needed to complete the coordination task.

\subsection{Task Taxonomy}

We organize tasks based on how much context an agent needs to solve them. In AEC workflows, difficulty is largely driven by whether the task can be solved from a single sheet, requires navigating across multiple sheets, or depends on coordinating information across different documents. This taxonomy provides a simple way to group tasks by scope while reflecting how real project coordination work is performed.

\textbf{Intra-Sheet:} Tasks that can be completed using a single sheet (one PDF page). These include checking whether callouts match the elements to which they point, verifying detail titles, or reviewing a local assembly. The focus is on understanding what is present on the page and correctly interpreting relationships between text and multimodal drawing elements.

\textbf{Intra-Drawing:} Tasks that require reasoning across multiple sheets within the same drawing set. Examples include validating cross-references, comparing sheet indices, and tracing details across views (2D cross-sections of 3D building models). These tasks require navigating between pages, interpreting and storing multimodal information, and keeping track of related information across the set.

\textbf{Intra-Project:} Tasks that involve multiple documents, such as drawings, specifications, and submittals. These include identifying conflicts between specifications and drawings, or evaluating compliance across sources. These tasks reflect real project-level coordination, where relevant information is distributed across different documents.

\subsection{Task Formulation}
Each AEC-Bench task instance consists of a natural-language instruction, a sandboxed execution environment, and an automated verifier. The environment contains real construction documents (e.g., drawings, specifications, or submittals) sourced from public-sector projects, along with pre-installed utilities for PDF rendering and text extraction. Tasks are defined using the Harbor task format and executed within the Harbor harness \cite{Harbor_Framework_Team_Harbor_A_framework_2026}, which provides a consistent interface for agent interaction and supports tool-based execution through terminal-style commands.
Given an instruction and environment, the agent must explore the document set, retrieve relevant information, and produce structured findings in a standardized JSONL output file. The framework is outcome-driven: agents are not evaluated on their intermediate actions or tool usage, but solely on the correctness and completeness of their final output as graded by a professional engineer or architect.
Each instance is scored using a task-specific automated verifier against known ground truth. Full credit is assigned for complete and correct findings, partial credit for partially correct outputs, and zero credit for incorrect or unsupported results.

\subsection{AEC-Bench Multimodal Subset}

\begin{figure*}[t]
\centering
\begin{subfigure}[t]{0.48\textwidth}
    \centering
    \includegraphics[width=\linewidth]{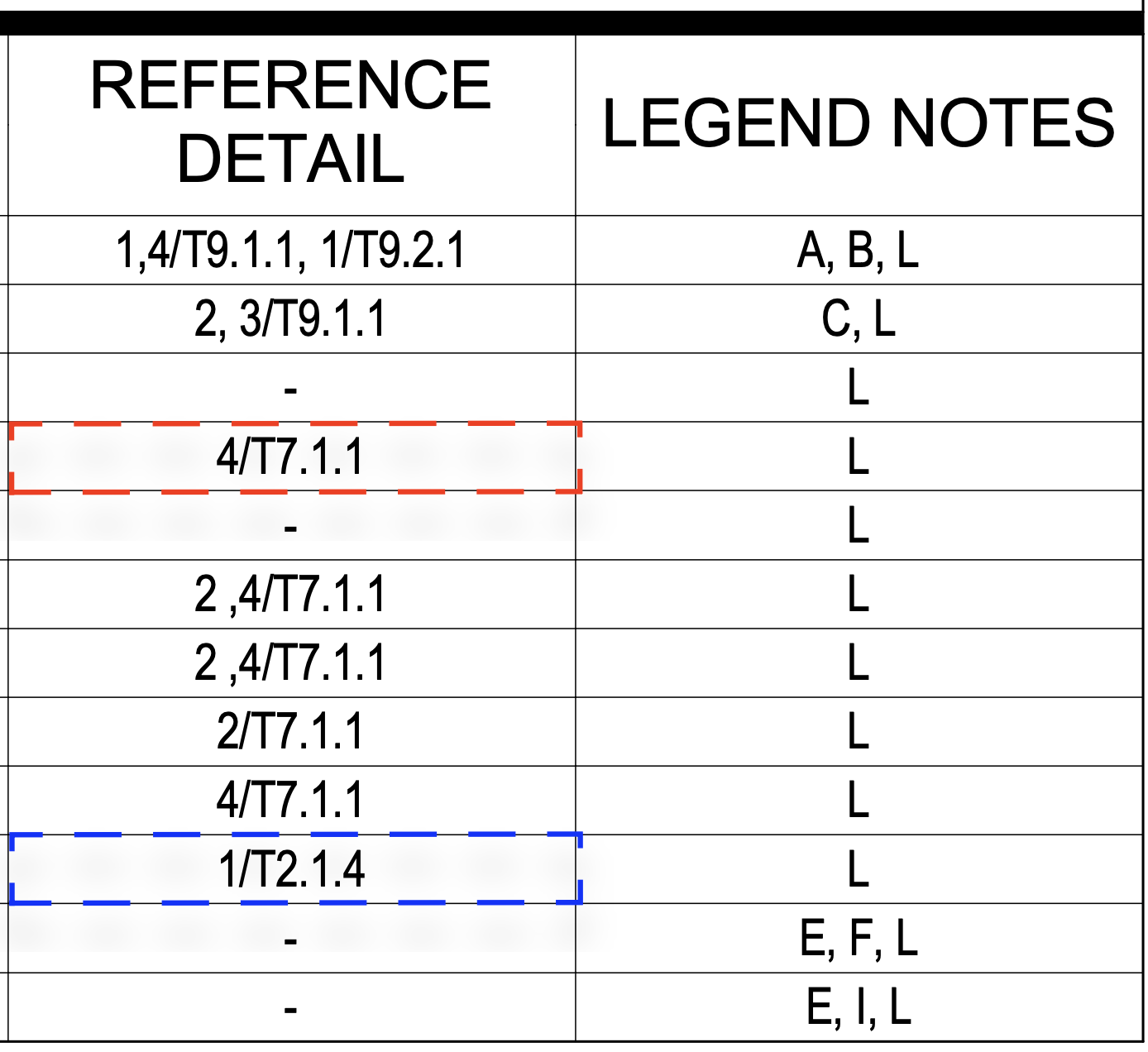}
    \caption{Cross-reference resolution \textbf{(before)}.}
\end{subfigure}
\hfill
\begin{subfigure}[t]{0.48\textwidth}
    \centering
    \includegraphics[width=\linewidth]{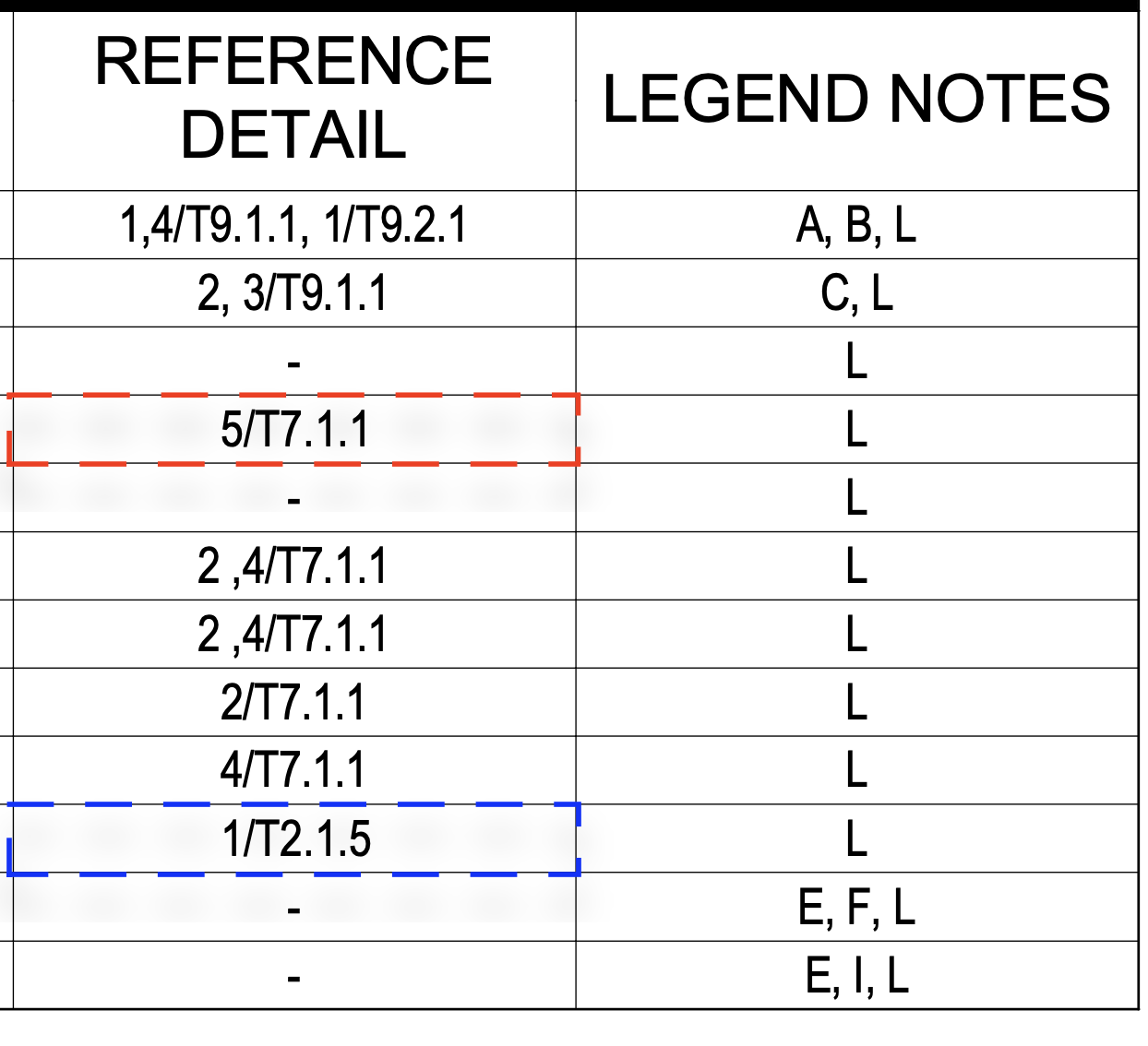}
    \caption{Cross-reference resolution \textbf{(after)}.}
\end{subfigure}

\vspace{0.5em}

\begin{subfigure}[t]{0.48\textwidth}
    \centering
    \includegraphics[width=\linewidth]{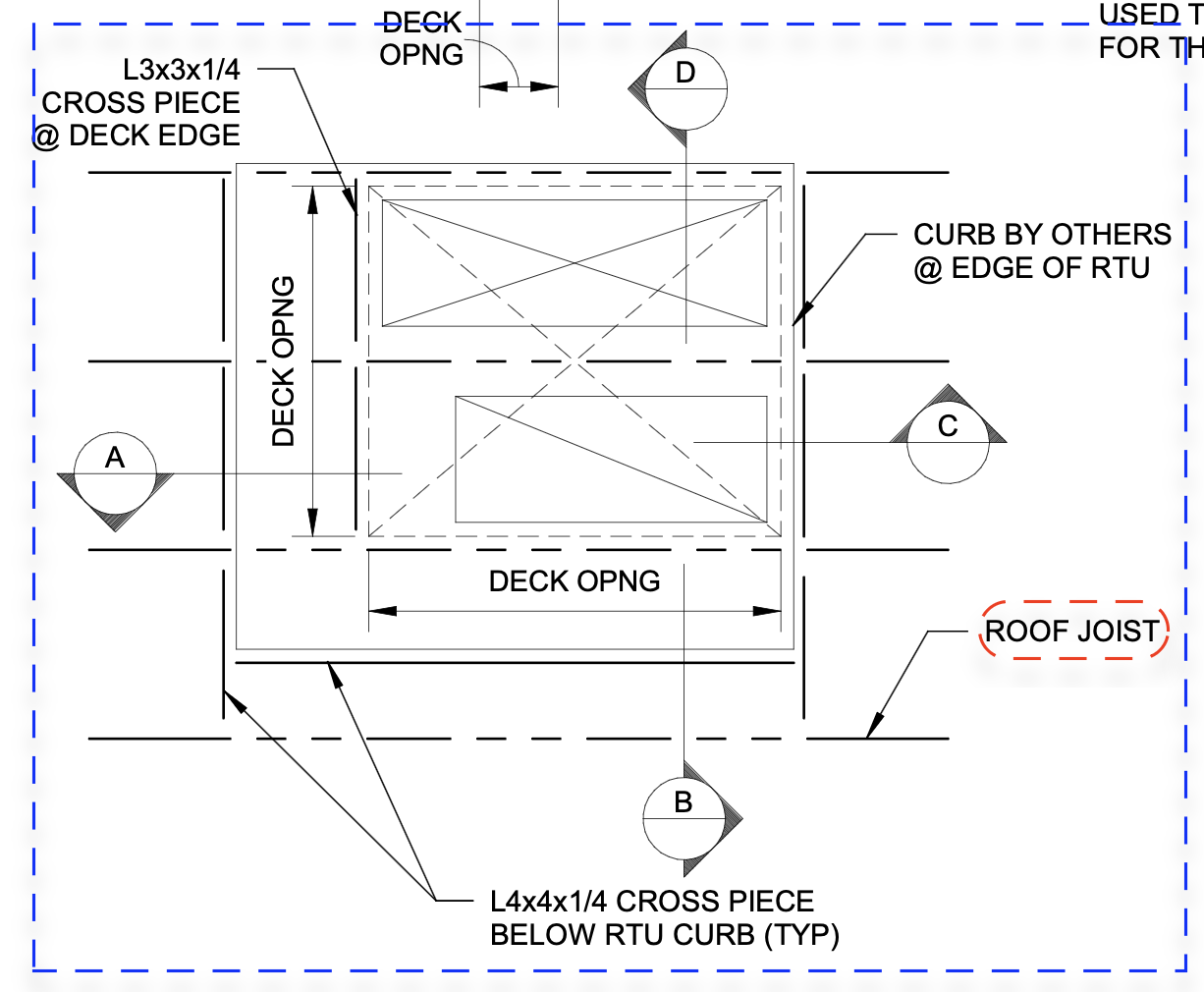}
    \caption{Note-callout accuracy \textbf{(before)}.}
\end{subfigure}
\hfill
\begin{subfigure}[t]{0.48\textwidth}
    \centering
    \includegraphics[width=\linewidth]{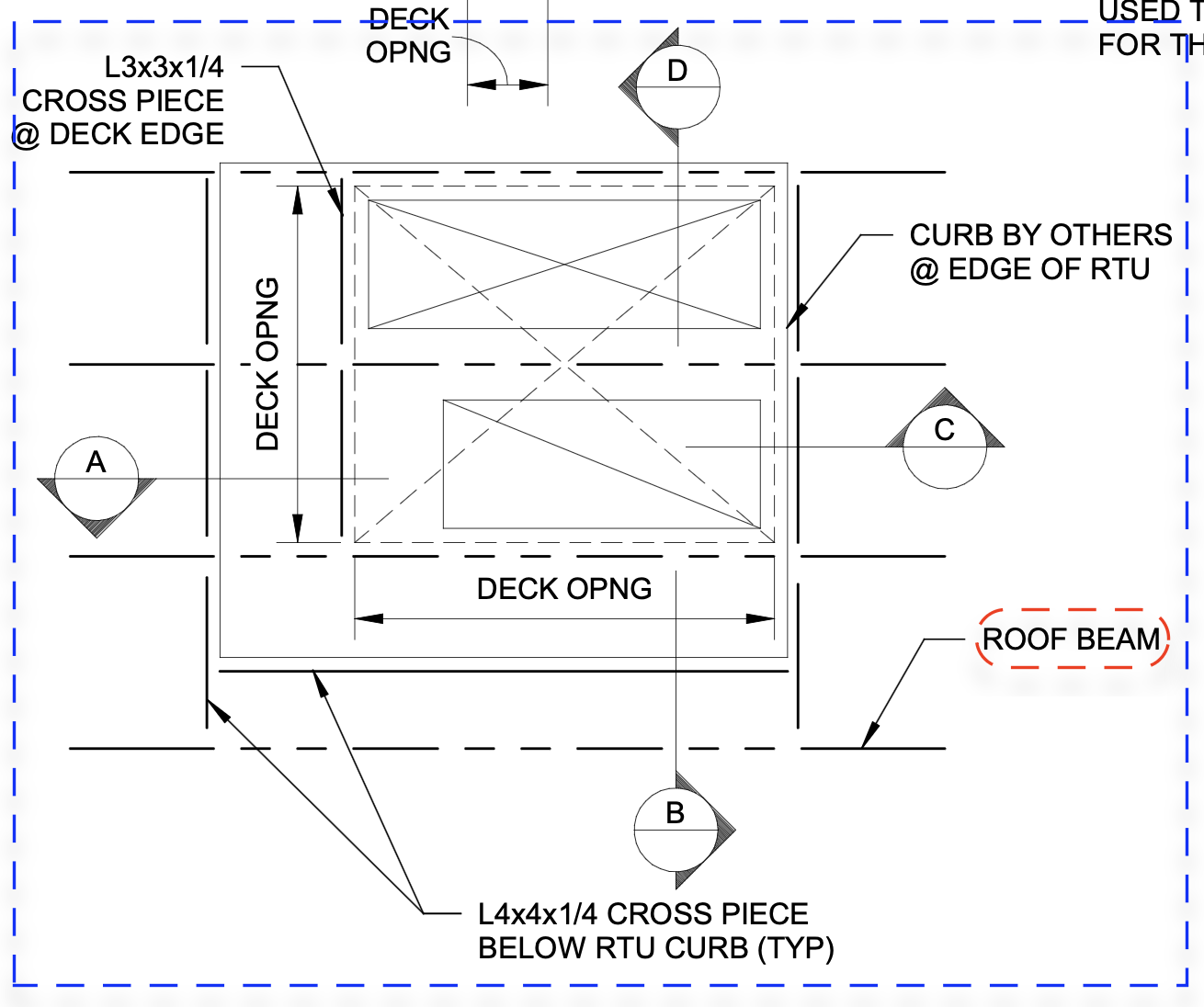}
    \caption{Note-callout accuracy task snapshot \textbf{(after)}.}
\end{subfigure}
\caption{Representative before/after task snapshots used in Section~2.2 data preparation.}
\label{fig:task-snapshots}
\end{figure*}

We construct AEC-Bench with a semi-automated pipeline that combines expert-authored task templates with structured extraction from real-world drawing packages sourced from publicly available PDF documents on the web, spanning multiple disciplines, including architectural, structural, civil, mechanical, electrical, and plumbing. Our PDF toolchain breaks these multi-page documents into structured, machine-readable data, including text with layout information, geometric regions, and cross-sheet references. We cache these artifacts with source coordinates for full traceability. Domain experts then select target pages and regions from this cache and inject realistic, precisely localized artifacts (for example, mismatched callout labels, broken cross-reference targets, and swapped specification values) using alignment-aware text editing that preserves visual fidelity. Each injected artifact is then verified with text-level assertions and pixel-level differencing (before, after, and diff) to confirm that the edit is structurally sound and visually consistent.
Figure~\ref{fig:task-snapshots} shows before/after examples for two task families: cross-reference resolution (top row) and note-callout accuracy (bottom row). In each pair, the left panel is the original sheet region, and the right panel is the edited version with a controlled injected artifact. For cross-reference resolution, the edits introduce subtle reference-number inconsistencies that require cross-sheet verification; for note-callout accuracy, the edit changes callout text while preserving surrounding geometry and leader structure to test precise text-to-geometry grounding. In general, the snapshots illustrate how evaluation artifacts are introduced with minimal visual disruption to maintain a realistic drawing context. The final subset contains 196 task instances in nine task types and three scopes. Table~\ref{tab:aecbench-dataset-prep} summarizes each task with a concise description and instance counts.

\begin{table*}[t]
\centering
\caption{AEC-Bench dataset summary by task. The benchmark is organized into three categories based on context scope: \textbf{Intra-Sheet} (single-page reasoning), \textbf{Intra-Drawing} (multi-sheet reasoning within a drawing set), and \textbf{Intra-Project} (cross-document reasoning across drawings, specifications, and submittals). The \textbf{Instances} column reports the number of task instances used for evaluation. }
\label{tab:aecbench-dataset-prep}
\small
\setlength{\tabcolsep}{4pt}
\renewcommand{\arraystretch}{1.1}
\begin{tabularx}{\textwidth}{@{}l l >{\raggedright\arraybackslash}X c@{}}
\toprule
\textbf{Category} & \textbf{Task} & \textbf{Description} & \textbf{Instances} \\
\midrule

\textbf{Intra-Sheet}
& detail-technical-review 
& Answer localized technical questions about details. 
& 14 \\

& detail-title-accuracy 
& Verify whether detail titles match the drawn content. 
& 15 \\

& note-callout-accuracy 
& Verify whether callout text correctly describes the referenced element. 
& 14 \\

\midrule

\textbf{Intra-Drawing}
& cross-reference-resolution 
& Identify cross-references that do not resolve to valid targets. 
& 51 \\

& cross-reference-tracing 
& Find all source locations referencing a given target detail. 
& 24 \\

& sheet-index-consistency 
& Compare sheet index entries against title blocks for mismatches. 
& 14 \\

\midrule

\textbf{Intra-Project}
& drawing-navigation 
& Locate the correct file, sheet, and detail given a query. 
& 12 \\

& spec-drawing-sync 
& Identify conflicts between specifications and drawings. 
& 16 \\

& submittal-review 
& Evaluate submittals for compliance with specs and drawings. 
& 36 \\

\midrule
\textbf{Total} &  &  & \textbf{196} \\
\bottomrule
\end{tabularx}
\end{table*}

\begin{table*}[t]
\centering
\caption{AEC-Bench mean reward by task, model, and setup. Rows are grouped by benchmark category and task; each model has two columns: \textbf{H} (base harness) and \textbf{H+\nomicN } (base harness + Nomic Tools). Boldface marks the higher score for each model-task pair.}
\label{tab:aec-bench-results}
\small
\setlength{\tabcolsep}{4pt}
\begin{tabularx}{\textwidth}{@{}l>{\raggedright\arraybackslash}X*{8}{c}@{}}
\toprule
\textbf{Category} & \textbf{Task} 
& \multicolumn{2}{c}{\textbf{GPT-5.4}} 
& \multicolumn{2}{c}{\textbf{GPT-5.2}} 
& \multicolumn{2}{c}{\textbf{Opus 4.6}} 
& \multicolumn{2}{c}{\textbf{Sonnet 4.6}} \\
\cmidrule(lr){3-4} \cmidrule(lr){5-6} \cmidrule(lr){7-8} \cmidrule(lr){9-10}
& 
& \textbf{H} & \textbf{H+\nomicN}
& \textbf{H} & \textbf{H+\nomicN}
& \textbf{H} & \textbf{H+\nomicN}
& \textbf{H} & \textbf{H+\nomicN} \\
\midrule

\textbf{Intra-Sheet}
& detail-technical-review
& 35.7 & \textbf{71.4} & 60.7 & \textbf{85.7} & 35.7 & \textbf{78.6} & 53.6 & \textbf{78.6} \\
& detail-title-accuracy
& \textbf{60.0} & \textbf{60.0} & \textbf{60.0} & 40.0 & 46.7 & \textbf{73.3} & \textbf{86.7} & 73.3 \\
& note-callout-accuracy
& \textbf{28.6} & \textbf{28.6} & \textbf{28.6} & 14.3 & 0.0 & \textbf{35.7} & \textbf{42.9} & 35.7 \\

\midrule

\textbf{Intra-Drawing}
& cross-reference-resolution
& \textbf{84.3} & 77.5 & 61.0 & \textbf{67.6} & \textbf{79.0} & 72.5 & \textbf{73.9} & 68.6 \\
& sheet-index-consistency
& \textbf{97.6} & 81.9 & 82.1 & \textbf{85.0} & 71.4 & \textbf{85.5} & 72.6 & \textbf{76.0} \\
& cross-reference-tracing
& \textbf{89.2} & 77.1 & \textbf{79.5} & 73.8 & 56.4 & \textbf{62.0} & 59.2 & \textbf{69.3} \\

\midrule

\textbf{Intra-Project}
& spec-drawing-sync
& 55.0 & \textbf{71.8} & 44.0 & \textbf{50.0} & 29.0 & \textbf{51.3} & 26.0 & \textbf{64.1} \\
& drawing-navigation
& 66.7 & \textbf{100.0} & \textbf{83.3} & \textbf{83.3} & 75.0 & \textbf{100.0} & 75.0 & \textbf{91.7} \\
& submittal-review
& 15.0 & \textbf{19.0} & 11.8 & \textbf{19.0} & \textbf{17.1} & 16.7 & 6.5 & \textbf{23.1} \\

\bottomrule
\end{tabularx}
\end{table*}

\section{Baseline Agent Evaluation Set-up}
We evaluate agent performance using baseline harness configurations designed to isolate the impact of tool access, document representation, and domain-specific augmentation. Rather than comparing foundation models in isolation, our goal is to understand how harness design shapes performance on AEC tasks. We consider two general-purpose coding-agent harness families: Codex and Claude Code. In the base setting (\textbf{H}), agents operate in a sandboxed environment with terminal (Bash) access and standard utilities for navigating document space, including cli based PDF tools. Agents are also free to write and execute their own code to process documents, create intermediate cache files, and perform operations such as searching across multiple files. This setup reflects typical coding-agent workflows, where documents are treated as files to be searched, parsed, and manipulated through command-line tools, often supplemented with rendered images for visual inspection.

To evaluate the effect of structured representations, we augment this setup with domain-specific Nomic tools (\textbf{H+\nomicN}). In this configuration, agents are provided with structured representations of drawings, including extracted text, layout elements, and reference relationships-generated using tools like Nomic Parse and Nomic Embeddings \cite{nussbaum2025nomicembedtrainingreproducible}. This allows us to measure how improved access to multimodal structure affects performance without changing the underlying harness.

This evaluation design enables us to test two key hypotheses: whether the general-purpose coding-agent harnesses transfer to AEC tasks, whether structured parsing improves performance, and whether domain-aware orchestration can overcome the limitations of general-purpose systems. By holding the environment and tasks constant, we isolate the effect of representation and orchestration on agent performance.

\section{Results}
We report results across tool ablations of our AEC agent harness.
Table~\ref{tab:aec-bench-results} reports the mean reward (0-100; higher is better) by task in the three benchmark context categories (Intra-Sheet, Intra-Drawing, Intra-Project). Our reward uses task-specific accuracy metrics, where each instance is scored based on the correctness of the structured JSON output produced by the agent. For each task instance, we run a single trial and compute the reward directly from the verifier, without averaging over multiple runs. Depending on task complexity and context category, each instance may involve one or more documents, requiring agents to operate over varying context scopes. Reward functions are designed to evaluate multiple findings in a single task instance, capturing both the correctness and completeness of the agent’s output for acceptance by a trained architect and engineer.
Our baseline evaluation focus is on identifying the boundary points of two general purpose coding-agent harness families: Codex and Claude Code, executed in the same sandboxed environments and evaluated under identical output contracts. For each model, the table shows two conditions: \textbf{H}, the base harness without Nomic specific tools, and \textbf{H+\nomicN}, the same harness augmented with Nomic tools. This side-by-side layout isolates the effect of parse augmentation while keeping the underlying harness fixed.

\section{Discussion}
In this section, we analyze how agent performance is shaped by the tools, data, and environment provided to them. The reported scores reflect what each model can do within this setup, not what it could achieve in isolation. In real AEC workflows, this setup is always part of the system, and performance depends not just on reasoning but also on how effectively the agent can find, access, and verify the right information.

A central finding of the benchmark is that multi-modal AEC tasks tightly couple retrieval and reasoning, with retrieval frequently acting as the primary bottleneck. Agents often fail before reaching the core reasoning step because they cannot reliably locate the relevant sheet, detail, or document; once the correct context is retrieved, performance improves substantially. Table~\ref{tab:aec-bench-results} provides consistent evidence for this pattern. Under the \textbf{H+\nomicN}, we observe substantial average gains across models, specifically on retrieval-sensitive tasks. In the detail-technical-review, performance improves by an average of \textbf{+32.2\%} points across models. Similar gains are observed for spec-drawing-sync \textbf{+20.8\%} and drawing-navigation \textbf{+18.75\%}, with improvements holding consistently across foundation model families.  Notably, these tasks are characterized by a primary dependence on locating the correct sheet, detail, or document before reasoning can proceed. Improved access to structured or retrievable context with parsing models directly translates into higher task performance on retrieval-sensitive tasks.

However, the comparison across setups also shows that document parsing provides targeted benefits rather than uniform gains. On tasks that rely more heavily on visual-spatial grounding. For example, note-callout-accuracy decreases on average by $-3.6\%$ points across models, while cross-reference-resolution drops by approximately $-2.4\%$ points, sheet-index-consistency by $-0.1\%$ points, and cross-reference-tracing shows an average degradation of $-0.53\%$, indicating that parse does not substantially resolve the need for exhaustive traversal. The tasks note-callout-accuracy and cross-reference-tracing share a common characteristic: they require precise localization, alignment between text and geometry. As illustrated in Figure. \ref{fig:task-snapshots}(c,d), the model must trace leader lines and related geometry purely visually to correctly identify and report issues; this remains a concrete failure mode for current foundation model harnesses. Even when relevant evidence might exist from the parsing, agents frequently fail to localize it accurately. Hence, in such settings, adding parsed PDFs representations does not directly address the underlying gap and can instead introduce additional context that the agent must navigate. The increase in context is further reflected in token usage. Taken together, these results indicate that while parsing improves retrieval in text-dominant tasks, it is less effective for tasks requiring fine-grained visual grounding. 

Finally, submittal-review also highlights a distinct failure mode that differs from both retrieval- and grounding-dominated tasks. This results in consistently low performance across all models and setups (best reward 23.1), even when a parse is available. While the parse improves access to relevant text, it does not resolve the need for higher-level judgment or reduce the search space sufficiently in long trajectories. As a result, agents tend to over-generate findings, leading to a high rate of false positives. Additionally, submittal-review introduces a degree of subjectivity that is absent in more deterministic tasks. Correct outputs depend not only on retrieving evidence but also on applying domain-specific judgment and prioritization consistent with professional review standards. This makes evaluation inherently more sensitive to human interpretation and increases the likelihood of disagreement with the verifier. 

Another pattern emerges in tool call execution across agent harnesses. Despite operating on multimodal construction documents, models consistently default to a coding-oriented tool repertoire. Codex-based agents (GPT-5.2/5.4) rely entirely on Bash (100\% of interactions), executing every action as a shell command and avoiding higher-level abstractions such as structured read/write tools, effectively treating AEC documents as source code. Claude-based agents exhibit slightly more diversity (53\% Bash, 35\% Read), but still operate predominantly through CLI-style interactions. Across all models, 77\% of trajectories invoke \texttt{pdftotext}, indicating strong convergence toward a \texttt{pdftotext} extraction pipeline; while this enables efficient keyword-based search, it collapses spatial layout and geometric relationships into linear text, discarding critical visual structure. Codex agents also rely heavily on rasterization via \texttt{pdftoppm} (96\% of runs), whereas Claude agents use it far less frequently (32\% of runs). Yet, this increased use of image rendering does not lead to better performance on visually grounded tasks. This suggests that access to images alone is insufficient without the ability to interpret spatial relationships. Taken together, these patterns indicate that coding-agents use their existing interaction paradigm-command-line search, text extraction, and image rendering-rather than adapting to the multimodal structure of construction documents, leading to systematic failures in tasks that require geometric reasoning and precise spatial grounding.

This behavior becomes particularly clear on note-callout-accuracy when accounting for all 14 instances across 4 models (56 total runs), which separate into three categories: text-catchable cases (2 instances), visual-required cases (10 instances), and clean cases (2 instances). Restricting the analysis to defect-detection cases reveals a sharp performance divergence. Text-catchable instances achieve near-perfect performance (mean reward 100\%), while visual-required instances achieve only 5\%, with meaningful outputs observed primarily from Claude Sonnet and no success from other models. This large gap highlights a consistent failure mode: tasks that depend on tracing and geometric interpretation remain challenging for current systems, even under identical task structures and tool availability. Across models, failures are not due to a lack of access to visual information-trajectories show extensive image rendering and inspection-but rather an inability to translate that visual input into structured, spatially grounded judgments. Model behavior like this further reinforces the boundary point of their usage in such environments.

\subsection{Limitations}

The AEC-Bench has several limitations. The current benchmark includes a limited number of documents and a subset of tasks per category and AEC discipline, which may constrain both statistical robustness and the coverage of model capabilities. Additionally, tasks do not fully capture the breadth of abilities exhibited by modern LLM harnesses, and some task families rely on curated artifacts or controlled defects that may not reflect the full diversity of real-world AEC drawing sets. Most tasks also rely on deterministic evaluation procedures that check for specific keywords or structured outputs; while this enables scalable and reproducible evaluation, it may not fully capture nuanced human judgments of engineering correctness or practical relevance. Nevertheless, the benchmark provides a meaningful and representative framework for evaluating agent performance within AEC multimodal coordination workflows.

\subsection{Replicability and License}
You can access the benchmark, agent harnesses, and evaluation code at \href{https://github.com/nomic-ai/aec-bench}{https://github.com/Nomic-ai/aec-bench}. We release this domain-expert annotated data and code under an open-source Apache 2 license to promote the progress of agentic capability research across the built environment.

\subsection{Future Directions}

Future work should expand the scale and diversity of the benchmark to include larger drawing sets, more disciplines, and additional task families. Improving evaluation methods is another important direction, including verifiers that can assess evidence grounding and reasoning steps rather than relying solely on unidirectional static matching. Another promising direction is the development of agentic systems designed specifically for document navigation. Such systems could iteratively explore drawing sets, maintain spatial memory on sheets, and retrieve evidence regions prior to reasoning. Finally, incorporating stronger domain knowledge into multimodal models may improve performance on tasks that require engineering judgment rather than simple visual extraction.

\section{Conclusion}
AEC-Bench introduces a benchmark for evaluating multimodal, agentic reasoning in construction-document coordination workflows grounded in real-world design and engineering practices. By framing tasks around common review activities, such as reference tracing, navigation, and coordination across drawings and specifications, it reflects the structure and demands of practical AEC coordination workflows, with outputs graded by domain-experts.

This benchmark and the associated evaluation results indicate that multimodal data representation and context engineering play a central role in shaping AEC agent performance within current agent harnesses. Coding-agent systems transfer meaningfully to this domain, particularly for tasks that can be addressed through retrieval and structured reasoning. However, performance becomes less consistent in settings that require spatial grounding or domain-sensitive judgment, highlighting the importance of aligning agent interaction strategies with the structure of the underlying multimodal documents.

We further observe that while both visual inputs and structured document parsing improve access to relevant information, neither is sufficient in isolation. Effective performance emerges from how these modalities are coordinated and surfaced within the agent loop, rather than from any single representation. More broadly, these findings point to the importance of domain-aware system design, where multiple capabilities are orchestrated to match the demands of the task.

As agent benchmarks evolve, incorporating realistic document structures and workflows will be critical for understanding agent behavior in applied settings. We hope AEC-Bench serves as a useful step in this direction, enabling a more grounded evaluation of agent capabilities in multimodal document environments.

\bibliographystyle{acl_natbib}
\bibliography{aecbench}

\end{document}